\documentclass[a4paper]{article}
\usepackage{CNVSRC2023}
\usepackage{epsfig,amssymb,amsmath,booktabs}
\ninept
\pdfoutput=1
\title{The GUA-Speech System Description for CNVSRC Challenge 2023}

\name{Shengqiang Li, Chao Lei, Baozhong Ma, Binbin Zhang, Fuping Pan}

\address{General User Agent, Shanghai, China \\
{\small \tt \{shengqiang.li, chao.lei, baozhong.ma, binbin.zhang, fuping.pan\}@guasemi.com}}%
\begin{document}
\maketitle

\begin{abstract}
This study describes our system for Task 1 Single-speaker Visual Speech Recognition (VSR) \emph{fixed track} in the Chinese Continuous Visual Speech Recognition Challenge (CNVSRC) 2023. Specifically, we use intermediate connectionist temporal classification (Inter CTC) residual modules to relax the conditional independence assumption of CTC in our model. Then we use a bi-transformer decoder to enable the model to capture both past and future contextual information. In addition, we use Chinese characters as the modeling units to improve the recognition accuracy of our model. Finally, we use a recurrent neural network language model (RNNLM) for shallow fusion in the inference stage. Experiments show that our system achieves a character error rate (CER) of 38.09\% on the Eval set which reaches a relative CER reduction of 21.63\% over the official baseline,  and obtains a second place in the challenge.
\end{abstract}

\section{Data}
\label{sec:data}
In this section, we provide a complete description of the data profile used to model training.
Specifically, for Task 1 Single-speaker VSR \emph{fixed track}, only \emph{CN-CVS} \cite{chen2023cn} and \emph{CNVSRC-Single.Dev} is used to perform system development. The train set of \emph{CN-CVS} contains about 252 hours of video (175,058 utterances). The validation set of \emph{CN-CVS} contains about 1.4 hours of video (913 utterances). The test set of \emph{CN-CVS} contains about 1.4 hours of video (903 utterances). The train set of \emph{CNVSRC-Single.Dev} contains about 90.6 hours of video (25,038 utterances). The validation set of \emph{CNVSRC-Single.Dev} contains about 2 hours of video (568 utterances).

\section{System}
 \begin{figure}[t]
\begin{minipage}[b]{1.0\linewidth}
  \centering
  \centerline{\includegraphics[width=8cm]{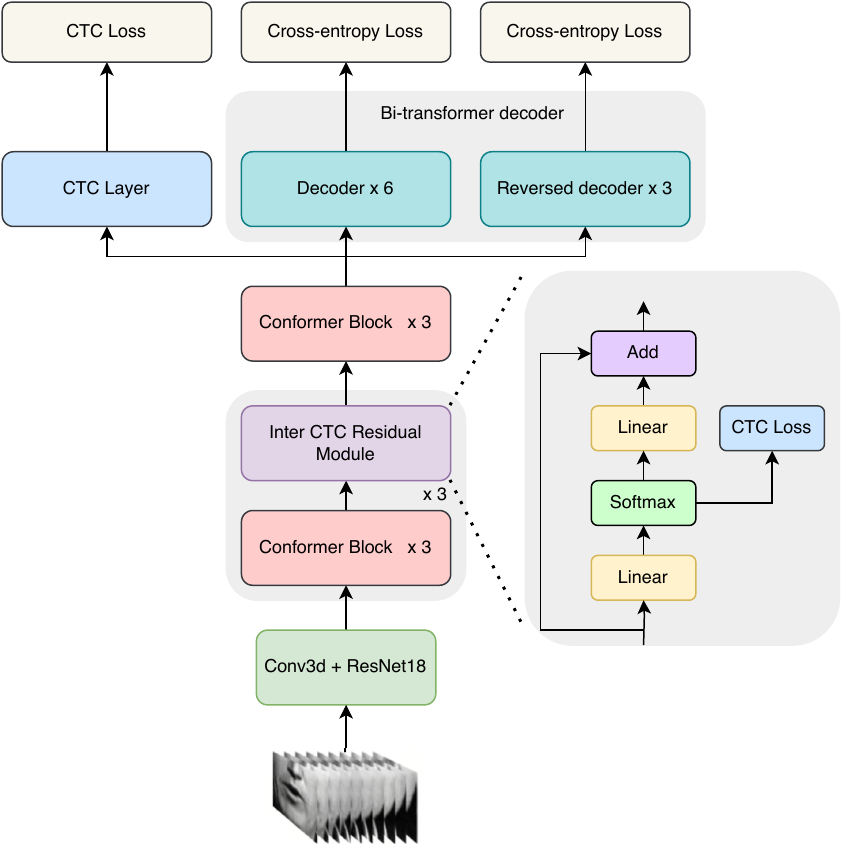}}
\end{minipage}
\caption{Architecture of the proposed VSR model. Intermediate predictions in the Inter CTC residual module are summed to the input of the next Conformer block to condition the prediction of the final block on it. Intermediate CTC losses are added to the output CTC loss for the computation of the final loss.}
\label{fig:model}
\end{figure}

Figure \ref{fig:model} shows the proposed VSR model, which consists of a visual front-end, 12 conformer encoder blocks with 3 Inter CTC residual modules, a bi-transformer decoder, and a CTC layer. In this section, we will provide a complete description of the model structure along with their vital configurations.
\subsection{Visual Front-end}
The visual front-end network \cite{ma2021end} transforms input video frames into temporal sequences. A 3D convolution stem with kernel size 5 × 7 × 7 is first applied to the video. Each video frame is then processed independently using a 2D ResNet-18 \cite{he2016deep} with an output spatial average pooling. Temporal features are then projected to the back-end network input dimension using a linear layer.
\subsection{Back-end Networks}
The back-end networks use a Conformer architecture \cite{ma2021end, gulati2020conformer} as the building blocks. In addition, Inter CTC residual modules and a Bi-transformer decoder are used to improve the performance of the model.
\subsubsection{Intermediate CTC residual module}
We add Inter CTC residual modules \cite{lee2021intermediate, nozaki2021relaxing, burchi2023audio} in encoder networks. We condition intermediate block features of visual encoders on early predictions to relax the conditional independence assumption of CTC models. During training and inference, each intermediate prediction is summed to the input of the next layer to help recognition. The output feature of the $l$-th encoder block $\boldsymbol{X}^l_{\text {out }}$ is passed through a feed-forward network with residual connection and a softmax activation function: 
\begin{equation}
\begin{aligned}
& \boldsymbol{Z}_l=\operatorname{Softmax}\left(\operatorname { Linear }\left(\boldsymbol{X}^l_{\text {out }}\right)\right) \\
& \boldsymbol{X}^{l+1}_{\text {in }}=\boldsymbol{X}^l_{\text {out }}+\operatorname{Linear}\left(\boldsymbol{Z}_l\right)
\end{aligned}
\end{equation}
Where $\boldsymbol{X}^{l + 1}_{\text {in}}$ is the input feature of the next encoder block, $\boldsymbol{Z}_l \in \mathbb{R}^ {T \times V}$ is a probability distribution over the
output vocabulary. The intermediate CTC loss of $k$-th Inter CTC residual module $\mathcal{L}^k_{\text {inter }}$ is then computed using the target sequence $\boldsymbol{y}$ as:
\begin{equation}
\begin{aligned}
& \mathcal{L}^k_{\text {inter }}=-\log \left(p\left(\boldsymbol{y} \mid \boldsymbol{Z}^l\right)\right) \\
& \text { with } p\left(\boldsymbol{y} \mid \boldsymbol{Z}^l\right)=\sum_{\pi \in \mathcal{B}_{C T C}^{-1}(\boldsymbol{y})} \prod_{t=1}^T \boldsymbol{Z}_{t, \pi_t}
\end{aligned}
\end{equation}
Where $\pi \in V^T$ are paths of tokens and $\mathcal{B}_{C T C}$ is a many-to-one map that simply removes all blanks and repeated labels.  The total intermediate CTC loss $\mathcal{L}_{\text {inter }}$ is computed as:
\begin{equation}
 \mathcal{L}_{\text {inter}} = \frac{1}{K}\sum_{k=1}^K \mathcal{L}^k_{\text {inter}}
\end{equation}
Where $K$ is the total number of the Inter CTC residual modules in the encoder. We use the Inter CTC residual module for every three conformer blocks with $K$ set to 3.
\subsubsection{Bi-transformer Decoder}
Inspired by our early work \cite{wu2021u2++, zhang2022wenet} on ASR task, we use the bi-transformer decoder to enable the model to capture both past and future contextual information. During inference, we use the left decoder only. During training, the loss function of the left decoder $\mathcal{L}_{left}$ is computed as:
\begin{equation}
\mathcal{L}_{left} = -\log \left(\prod_{l=1}^L p\left(y_l \mid \boldsymbol{y}_{1: l-1}, \boldsymbol{X}_{\text{e}}\right)\right)
\end{equation}
The loss function of the right decoder $\mathcal{L}_right$ is computed as:
\begin{equation}
\mathcal{L}_{right} = -\log \left(\prod_{l=L}^1 p\left(y_l \mid \boldsymbol{y}_{L: l-1}, \boldsymbol{X}_{\text{e}}\right)\right)
\end{equation}
Where $\boldsymbol{y}=\left(y_1, \cdots, y_L\right)$ denotes the target sequence, and $\boldsymbol{X}_{\text{e}}$ denotes the output of the encoder. The total loss function of the bi-transformer decoder is:
\begin{equation}
\mathcal{L}_{attn} = \left( 1 - \alpha \right) \mathcal{L}_{left} + \alpha \mathcal{L}_{right}
\end{equation}
Where $\alpha$ is a tunable hyper-parameter. In our system, $\alpha$ is set to 0.3.
\subsection{Objective Function}
The loss function of the proposed vsr model is computed as:
\begin{equation}
\mathcal{L} = \lambda \left( \gamma \mathcal{L}_{inter} + \left( 1 - \gamma \right) \mathcal{L}_{ctc} \right) + \left(1 - \lambda \right) \mathcal{L}_{attn}
\end{equation}
where $\mathcal{L}_{ctc}$ is an auxiliary CTC loss function \cite{graves2006connectionist}, and $\lambda$ and $\gamma$ are two tunable hyper-parameters. In our system, $\lambda$ is set to 0.1, and $\gamma$ is set to 0.3.
\section{Experiments}
\subsection{Pre-processing}
In this section, we provide a complete description of the data processing pipeline. The RetinaFace \cite{deng2020retinaface} face detector and Face Alignment Network (FAN) \cite{bulat2017far} are used to detect 68 facial landmarks. Similar to \cite{ma2021end}, we remove differences related to rotation and scale by cropping the lip regions using bounding boxes of 96 × 96 pixels to facilitate recognition. The cropped images are then converted to gray-scale and normalized between $-1$ and $1$. In the training stage, video streams are augmented with random cropping and adaptive time masking \cite{ma2022visual}. We use Chinese characters as modeling units, the vocabulary consists of 4466 Chinese characters generated from the text of the train set of \emph{CN-CVS} and CNVSRC-Single.Dev, and 3 special tokens for $<blank>$, $<unk>$, $<eos/sos>$.
\subsection{Experimental settings}
All of our experiments were implemented using CNVSRC 2023 Baseline\footnote{https://github.com/MKT-Dataoceanai/CNVSRC2023Baseline}. Following the default set in CNVSRC 2023 Baseline, we use 12 conformer layers in the encoder where the attention dimension is 768, the number of attention heads is 12, the kernel size of the CNN module is 31, and the feed-forward network dimension is 3072. The bi-transformer decoder consists of 6 transformer decoders and 3 reversed transformer decoders, where the attention dimension is 768, the number of attention heads is 12 and the feed-forward network dimension is 3072. The RNN language model has two layers and the hidden size of each layer is 650.
\subsection{Training}
The vsr model is trained from scratch through curriculum learning. Firstly, we train the vsr model using the subset that includes only short utterances lasting no more than 4 seconds of \emph{CN-CVS} and average the model weights from epoch 14 to epoch 23. Secondly, we use the averaged checkpoint from stage 1 to initialize the vsr model and train the vsr model with the full dataset of \emph{CN-CVS} and average the model weights from epoch 65 to epoch 74. Finally, we use the averaged checkpoint from stage 2 to initialize the vsr model and train the vsr model with the train set of \emph{CNVSRC-Single.Dev}. For the main results shown in Table \ref{tab.main_result}, the model is trained using 25038 utterances of \emph{CNVSRC-Single.Dev} as the train set and 568 utterances of \emph{CNVSRC-Single.Dev} as the valid set, which is the same as mentioned in Section \ref{sec:data}. For the ablation study shown in Table \ref{tab.ablation_study}, the model is trained using 22767 utterances of \emph{CNVSRC-Single.Dev} as the train set and 2839 utterances of \emph{CNVSRC-Single.Dev} as the valid set, the same as the CNVSRC 2023 Baseline. The RNN language model is trained using the train set of \emph{CN-CVS} and \emph{CNVSRC-Single.Dev} for 60 epochs. 

\subsection{Inference}
Decoding is performed using joint CTC/attention one-pass decoding \cite{watanabe2017hybrid}, and an RNN language model is used for shallow fusion. During decoding, the CTC weight is 0.3, the lm weight is 0.1, and the beam size is 40.
\subsection{Results}
\subsubsection{Main Results}
Table \ref{tab.main_result} reports the character error rate (CER) of the proposed system on the eval set. From the table, we see that our system achieves a CER of 38.09\%, which reaches a relative CER reduction of 21.63\% over the official baseline.
\begin{table}[t]
    \setlength{\abovecaptionskip}{0pt}
    \setlength{\belowcaptionskip}{10pt}
    \caption{WER (\%) Comparison between the proposed system and the official baseline on the eval set.}
    \centering
        \scalebox{1.0}{
	\begin{tabular}{ccc}
            \toprule
		\textbf{System} & \textbf{Model} & CER (\%)\\
		\midrule
		B1 & Official baseline & 48.60 \\
		M1 & Proposed system  & 38.09 \\
            \bottomrule
	\end{tabular}}
	\label{tab.main_result}
\end{table}
\subsubsection{Ablation study}
Table \ref{tab.ablation_study} shows the ablation study of the proposed system on the valid set. Valid set 1 contains 2839 utterances, which is the same as the CVSRC2023 Baseline. Valid set 2 contains 568 utterances and is a subset of the valid set 1.  D1 denotes that the train set and the valid set are the same as the CVSRC2023 Baseline. D2 denotes that the train set and the valid set are the same as described in Section \ref{sec:data}. From the table, we see that the CER of the same model in valid set 1 is almost the same as the CER in valid set 2, which means that the valid set 2 is effective as valid set 1. Firstly, Inter CTC residual module and Bi-transformer decoder give a noticeable gain. Secondly, using Chinese characters as the modeling unit also improves the recognition accuracy. In addition, the RNN language model improves the performance of the system. Most importantly, the more utterances in \emph{CNVSRC-Single.Dev} used as the train set, the more accuracy the model achieves.
\begin{table}[t]
    \setlength{\abovecaptionskip}{0pt}
    \setlength{\belowcaptionskip}{10pt}
    \caption{WER (\%) Ablation study of the proposed system on the valid set. The evaluation metric is the character error rate (CER\%).}
    \centering
        \scalebox{0.8}{
	\begin{tabular}{cccc}
            \toprule
		\textbf{System} & \textbf{Model} & \textbf{Valid set 1} & \textbf{Valid set 2} \\
		\midrule
            M1D2 & Proposed system & - & 36.46 \\
            M1D1 & Proposed system & 40.46 & 40.37 \\
            M2D1 & M1D1 - RNNLM  & 40.62 & 40.51 \\
            M3D1 & M2D1 - char unit  & 42.36 & 42.38 \\
            M4D1 & M3D1 - Bi-transformer decoder  & 43.19 & 43.15 \\
            M5D1 & M4D1 - Inter CTC residual module  & 48.57 & 48.34 \\
            \bottomrule
	\end{tabular}}
	\label{tab.ablation_study}
\end{table}

\bibliographystyle{IEEEbib}
\bibliography{BibEntries}

\begin{thebibliography}{10}

\bibitem{chen2023cn}
Chen Chen, Dong Wang, and Thomas~Fang Zheng,
\newblock ``Cn-cvs: A mandarin audio-visual dataset for large vocabulary continuous visual to speech synthesis,''
\newblock in {\em ICASSP 2023-2023 IEEE International Conference on Acoustics, Speech and Signal Processing (ICASSP)}. IEEE, 2023, pp. 1--5.

\bibitem{ma2021end}
Pingchuan Ma, Stavros Petridis, and Maja Pantic,
\newblock ``End-to-end audio-visual speech recognition with conformers,''
\newblock in {\em ICASSP 2021-2021 IEEE International Conference on Acoustics, Speech and Signal Processing (ICASSP)}. IEEE, 2021, pp. 7613--7617.

\bibitem{he2016deep}
Kaiming He, Xiangyu Zhang, Shaoqing Ren, and Jian Sun,
\newblock ``Deep residual learning for image recognition,''
\newblock in {\em Proceedings of the IEEE conference on computer vision and pattern recognition}, 2016, pp. 770--778.

\bibitem{gulati2020conformer}
Anmol Gulati, James Qin, Chung-Cheng Chiu, Niki Parmar, Yu~Zhang, Jiahui Yu, Wei Han, Shibo Wang, Zhengdong Zhang, Yonghui Wu, et~al.,
\newblock ``Conformer: Convolution-augmented transformer for speech recognition,''
\newblock {\em arXiv preprint arXiv:2005.08100}, 2020.

\bibitem{lee2021intermediate}
Jaesong Lee and Shinji Watanabe,
\newblock ``Intermediate loss regularization for ctc-based speech recognition,''
\newblock in {\em ICASSP 2021-2021 IEEE International Conference on Acoustics, Speech and Signal Processing (ICASSP)}. IEEE, 2021, pp. 6224--6228.

\bibitem{nozaki2021relaxing}
Jumon Nozaki and Tatsuya Komatsu,
\newblock ``Relaxing the conditional independence assumption of ctc-based asr by conditioning on intermediate predictions,''
\newblock {\em arXiv preprint arXiv:2104.02724}, 2021.

\bibitem{burchi2023audio}
Maxime Burchi and Radu Timofte,
\newblock ``Audio-visual efficient conformer for robust speech recognition,''
\newblock in {\em Proceedings of the IEEE/CVF Winter Conference on Applications of Computer Vision}, 2023, pp. 2258--2267.

\bibitem{wu2021u2++}
Di~Wu, Binbin Zhang, Chao Yang, Zhendong Peng, Wenjing Xia, Xiaoyu Chen, and Xin Lei,
\newblock ``U2++: Unified two-pass bidirectional end-to-end model for speech recognition,''
\newblock {\em arXiv preprint arXiv:2106.05642}, 2021.

\bibitem{zhang2022wenet}
Binbin Zhang, Di~Wu, Zhendong Peng, Xingchen Song, Zhuoyuan Yao, Hang Lv, Lei Xie, Chao Yang, Fuping Pan, and Jianwei Niu,
\newblock ``Wenet 2.0: More productive end-to-end speech recognition toolkit,''
\newblock {\em arXiv preprint arXiv:2203.15455}, 2022.

\bibitem{graves2006connectionist}
Alex Graves, Santiago Fern{\'a}ndez, Faustino Gomez, and J{\"u}rgen Schmidhuber,
\newblock ``Connectionist temporal classification: labelling unsegmented sequence data with recurrent neural networks,''
\newblock in {\em Proceedings of the 23rd international conference on Machine learning}, 2006, pp. 369--376.

\bibitem{deng2020retinaface}
Jiankang Deng, Jia Guo, Evangelos Ververas, Irene Kotsia, and Stefanos Zafeiriou,
\newblock ``Retinaface: Single-shot multi-level face localisation in the wild,''
\newblock in {\em Proceedings of the IEEE/CVF conference on computer vision and pattern recognition}, 2020, pp. 5203--5212.

\bibitem{bulat2017far}
Adrian Bulat and Georgios Tzimiropoulos,
\newblock ``How far are we from solving the 2d \& 3d face alignment problem?(and a dataset of 230,000 3d facial landmarks),''
\newblock in {\em Proceedings of the IEEE international conference on computer vision}, 2017, pp. 1021--1030.

\bibitem{ma2022visual}
Pingchuan Ma, Stavros Petridis, and Maja Pantic,
\newblock ``Visual speech recognition for multiple languages in the wild,''
\newblock {\em Nature Machine Intelligence}, vol. 4, no. 11, pp. 930--939, 2022.

\bibitem{watanabe2017hybrid}
Shinji Watanabe, Takaaki Hori, Suyoun Kim, John~R Hershey, and Tomoki Hayashi,
\newblock ``Hybrid ctc/attention architecture for end-to-end speech recognition,''
\newblock {\em IEEE Journal of Selected Topics in Signal Processing}, vol. 11, no. 8, pp. 1240--1253, 2017.

\end{thebibliography}

\end{document}